\RequirePackage{silence}
\documentclass[sn-mathphys,Numbered]{sn-jnl}

\usepackage{lmodern}
\usepackage{anyfontsize}
\usepackage{graphicx}
\usepackage{multirow}
\usepackage{amsmath,amssymb,amsfonts}
\usepackage{amsthm}
\usepackage{mathrsfs}
\usepackage[title]{appendix}
\usepackage{xcolor}
\usepackage{textcomp}
\usepackage{manyfoot}
\usepackage{booktabs}
\usepackage{tabularx}
\usepackage{makecell}
\usepackage{array}
\usepackage{algorithm}
\usepackage{algorithmicx}
\usepackage{algpseudocode}
\usepackage{listings}
\usepackage{url}
\usepackage{rotating}

\begin{document}
\raggedbottom

\title[Learning When to Remember]{Learning When to Remember: Risk-Sensitive Contextual Bandits for Abstention-Aware Memory Retrieval in LLM-Based Coding Agents}

\author*[1]{\fnm{Mehmet} \sur{Iscan}}\email{miscan@yildiz.edu.tr}
\affil*[1]{\orgdiv{PythaLab}, \orgname{Yildiz Technical University}, \orgaddress{\city{Istanbul}, \country{T\"urkiye}}}

\abstract{Large language model (LLM)-based coding agents increasingly depend on external memory to reuse previous debugging experience, historical repair traces, and repository-local operational knowledge. This memory is valuable only when the current failure is genuinely compatible with a previous one. In software debugging, superficial similarity is often misleading: different root causes can expose similar stack traces, terminal errors, paths, or configuration symptoms. Consequently, unsafe memory injection can anchor the agent on an incorrect repair strategy, consume context budget, and amplify hallucinated fixes. This paper reframes issue-memory use as a selective, risk-sensitive control problem rather than a pure top-$k$ retrieval problem. We introduce \textbf{RSCB-MC}, a risk-sensitive contextual bandit memory controller that decides whether an agent should use no memory, inject the top resolution, summarize multiple candidates, perform high-precision or high-recall retrieval, abstain, or ask for feedback. The system stores reusable issue knowledge through a pattern-variant-episode schema and converts retrieval evidence into a fixed 16-feature contextual state that captures relevance, uncertainty, structural compatibility, feedback history, false-positive risk, latency, and token cost. Its reward design intentionally penalizes false-positive memory injection more strongly than missed reuse, making non-injection and abstention first-class safety actions. In the supplied deterministic smoke-scale artifacts, RSCB-MC obtains the strongest non-oracle offline replay success rate, \textbf{62.5\%}, while maintaining a \textbf{0.0\%} false-positive rate. In a bounded 200-case hot-path validation, it reaches \textbf{60.5\%} proxy success with \textbf{0.0\%} false positives and a \textbf{331.466 $\mu$s} p95 decision latency. The evidence is intentionally scoped: the current evaluation is local, deterministic, and proxy-based rather than a causal online LLM-agent deployment study. The central result is nevertheless clear: for coding-agent memory, the main question is not only which memory is most similar, but whether any retrieved memory is safe enough to influence the debugging trajectory.}

\maketitle


\section{Introduction}\label{sec:introduction}

This work began with a concrete operational frustration rather than with an abstract retrieval question. While operating an LLM-based coding agent on a working Python repository, we repeatedly observed the same pattern: the agent would recognize a familiar-looking error, reach for a previously stored ``fix,'' and confidently apply it---only to make the situation worse. A SQLite \texttt{database is locked} message and a stale migration produced almost identical surface logs, but the first required releasing a lingering connection while the second required re-running \texttt{alembic upgrade head}. A failure caused by activating the wrong virtual environment looked, on the terminal, indistinguishable from a wrong \texttt{PYTHONPATH}: both ended in \texttt{ModuleNotFoundError}, both pointed at the same file, and both had been ``solved'' by previous memory entries that were no longer applicable. An invalid configuration key for an internal service threw the same generic \texttt{KeyError} as a path-resolution failure for a backup directory. In each case, retrieval by surface similarity was not merely unhelpful; it actively pushed the agent down the wrong repair branch and consumed context budget on a misleading hint.

This is the practical mismatch that motivated RSCB-MC. LLM coding agents have moved beyond passive completion toward autonomous systems that inspect repositories, run tests, read traces, and iteratively repair failures \cite{m1-2,m1-3,m8-6}, and once such an agent can execute shell commands and persist across sessions, memory stops being optional prompt decoration and becomes a safety surface in its own right \cite{m1-1}. The agent has to remember which fixes were already attempted, which diagnostic paths failed, and which repository-specific lessons have been validated, because debugging sessions accumulate commands, stack traces, dependency versions, test outputs, hypotheses, and human feedback that no finite context window can hold \cite{m3-1,m3-4,m8-8}. Recent reinforcement-learning work on Memory-R1 and Agentic Memory shows that memory operations themselves can be learned rather than hard-coded \cite{m7-12,m7-14}. The question we faced was narrower and sharper: given that we already have such a memory, when should we be willing to actually \emph{use} it?

The broader RAG literature gives a partial answer but stops short of our setting. Knowledge-boundary work argues that retrieval is sometimes beneficial, sometimes neutral, and sometimes harmful, depending on the query and on what the model already knows \cite{m2-9}. Parametric RAG shows that simply adding more retrieved context raises online cost and can degrade reasoning when the evidence is poorly integrated \cite{m2-6}. Knowledge-base access methods such as KnowledGPT and direct retrieval-augmented optimization improve \emph{how} external knowledge is selected and represented \cite{m2-4,m2-7}. None of these directly ask whether retrieved \emph{operational memory}---a piece of historical debugging evidence about a specific repository---is safe to be inserted into an agent that will then run commands and edit files based on it.

The automated-debugging and APR literature confirms that prior fixes are valuable, but it implicitly inherits the same blind spot. History-driven program repair learns from recurring human fix patterns \cite{m4-7}; RAP-Gen retrieves bug-fix pairs to guide patch generation \cite{m4-8}; ReAPR injects retrieved bug-fix pairs into LLM repair prompts \cite{m4-1}; and Repair-Ingredients-Search separates root-cause ingredients from solution-phase ingredients \cite{m4-2}. These systems all assume that, once a sufficiently similar past fix is found, injecting it is the right move. The five concrete failure modes we encountered---SQLite lock vs.\ stale migration, wrong virtual environment vs.\ wrong \texttt{PYTHONPATH}, invalid configuration key vs.\ missing backup directory, retrieval false positive vs.\ rejected memory reuse, and outdated memory variant vs.\ migration-order mismatch---show why this assumption is unsafe in practice: each pair shares logs and command fragments while requiring a different fix.

Agentic RAG and process-supervised RL have begun to model retrieval as a trainable decision process. RAG-Gym formulates agentic RAG as a high-level Markov decision process and optimizes prompt engineering, actor tuning, and critic training \cite{m7-9}. ReasonRAG and TreePS-RAG argue that outcome-only rewards are too sparse and that process-level or tree-derived supervision improves multi-step retrieval behavior \cite{m7-8,m7-6}. SIRAG introduces lightweight agents that decide when to continue retrieval and which evidence to select \cite{m7-7}, and FLAIR adapts retrieval based on user feedback in a copilot setting \cite{m7-4}. These works are valuable because they move beyond static retrieve-then-generate pipelines, but their unit of decision is still ``which evidence improves the next answer.'' None of them treat false-positive \emph{memory injection in a debugging trajectory} as a first-class safety event with its own measurable cost.

The abstention and refusal literature provides the complementary lens. Selective prediction allows a model to abstain on low-confidence examples \cite{m5-1}; selective generation extends this to controllable language generation under risk control \cite{m5-3}; conformal abstention adapts thresholds for risk management \cite{m5-2}; independent rejectors improve reliability without repeatedly querying the LLM \cite{m5-4}; Learn-to-Refuse imposes a structured knowledge scope and refusal mechanism \cite{m5-6}; and I-CALM studies prompt-level incentives for confidence-aware abstention \cite{m5-8}. These methods establish that refusing or deferring is rational, but their target is almost always the final \emph{answer}. In our setting, the object being accepted or rejected is not an answer; it is a piece of retrieved operational memory that, if injected, will quietly shape every subsequent reasoning step the agent takes. A wrong final answer is visible to the user; a wrong memory injection is not.

These five observations---repeated false-positive injections in real debugging sessions, the gap between RAG safety and operational memory safety, the unsafe-by-default assumption in repair-ingredient retrieval, the answer-centric scope of agentic RAG, and the answer-centric scope of abstention---together pushed us toward a different formulation. We treat issue-memory use as \emph{risk-sensitive selective control}. Retrieval remains necessary, but its score is treated as evidence rather than permission. The controller observes a structured state built from candidate scores, ambiguity, failure-family compatibility, entity overlap, command/path/stack signatures, historical acceptance, historical false-positive rate, session rejection state, and operational cost. It then chooses whether to inject memory, compress memory, retrieve conservatively, abstain, ask for feedback, or proceed without memory. Two design choices follow directly from the failure modes above. First, because false-positive injection is the most damaging outcome we observed, the reward function makes its penalty strictly larger than the bonus for a correct injection. Second, because abstention was the action that would have saved us the most repair time in those sessions, abstention and \texttt{no\_memory} are first-class actions, not fallback states. The design is also aligned with context-budget work that treats context as a constrained resource \cite{m7-11} and with calibration-oriented RAG that argues relevance alone is insufficient when downstream decisions must be reliable \cite{m7-15}.

This paper introduces \textbf{RSCB-MC} (Risk-Sensitive Contextual Bandit with Memory Control). The method is intentionally lightweight: it does not fine-tune an LLM, does not require hosted inference, and does not perform expensive multi-trajectory rollouts in the benchmark pipeline. Instead, it operates as a local hot-path controller over a fixed action set. Its core contribution is the explicit modeling of unsafe memory reuse: false-positive memory injection is measured, penalized, and compared against safe non-injection behavior. The resulting question is simple but central to practical coding agents: \emph{when should the agent remember, and when should it deliberately not remember?}

The contributions are as follows. First, we define coding-agent issue-memory reuse as an abstention-aware, risk-sensitive contextual-bandit problem, motivated by the concrete failure modes encountered above. Second, we introduce a pattern-variant-episode memory representation that separates reusable root-cause patterns, context-specific variants, and validated episodes, so that two failures sharing a symptom but differing in root cause are not collapsed into one ``similar'' record. Third, we construct a fixed 16-feature state representation for auditable memory decisions. Fourth, we define a risk-sensitive reward and action score that penalize false-positive memory injection more strongly than missed reuse. Fifth, we evaluate the method through deterministic local artifacts covering retrieval, paraphrase robustness, hard negatives, abstention behavior, offline replay, reward sensitivity, ablations, context-budget proxies, and bounded hot-path validation. Finally, we position the method relative to RAG optimization, program-repair retrieval, memory-management RL, process-supervised agentic RAG, and abstention/refusal methods.

\section{Method}\label{sec:method}

\subsection{Task and notation}\label{subsec:task}

At time $t$, a coding agent observes a raw debugging context
\begin{equation}
    x_t \in \mathcal{X},
\end{equation}
where $x_t$ may contain an error message, stack excerpt, command, file path, repository scope, environment metadata, previous repair attempts, and session-level feedback. A deterministic normalization function maps this raw context to a query profile
\begin{equation}
    q_t = \phi(x_t).
\end{equation}
The profile $q_t$ contains normalized lexical evidence, root-cause tokens, failure-family signals, command signatures, path signatures, stack signatures, entity slots, and feedback-derived signals. A local memory bank $M$ stores prior issue knowledge. Retrieval produces a candidate set
\begin{equation}
    C_t = R(q_t,M)=\{c_{t,1},\ldots,c_{t,k}\}.
\end{equation}
A pure retrieval formulation would rank $C_t$ and inject the top candidate. RSCB-MC instead builds a policy state
\begin{equation}
    z_t = \psi(q_t,C_t)\in\mathbb{R}^{16}
\end{equation}
and chooses an action
\begin{equation}
    a_t \in A_t \subseteq A.
\end{equation}
The action may inject memory, request more information, or avoid memory entirely. The controller then observes feedback $y_t$, computes a reward $r_t$, and updates its policy state $\Theta$.

This is a contextual-bandit approximation of a richer debugging process. It deliberately optimizes the immediate memory-gating decision and does not claim to model all delayed multi-turn repair effects. That restriction is important: a production coding agent is a partially observable, non-stationary system, but the hot-path memory decision must still be fast enough to run before every memory injection.

\subsection{Pattern-variant-episode memory representation}\label{subsec:schema}

The memory bank is not treated as a flat vector store. Each reusable issue-memory record is organized into three levels:
\begin{equation}
    m_i = (p_i, V_i, E_i),
\end{equation}
where $p_i$ is a canonical pattern, $V_i$ is a set of variants, and $E_i$ is a set of observed episodes. A pattern stores a reusable symptom and root-cause class. A variant stores context-specific fix strategy and command/path/stack signatures. An episode stores concrete observed failure evidence, validation evidence, and feedback. This representation makes the memory more auditable than a single retrieved text chunk: the controller can inspect whether the current failure matches only the symptom, or also the operational signatures and previous validation context.

\begin{figure}[t]
\centering
\includegraphics[width=0.98\textwidth]{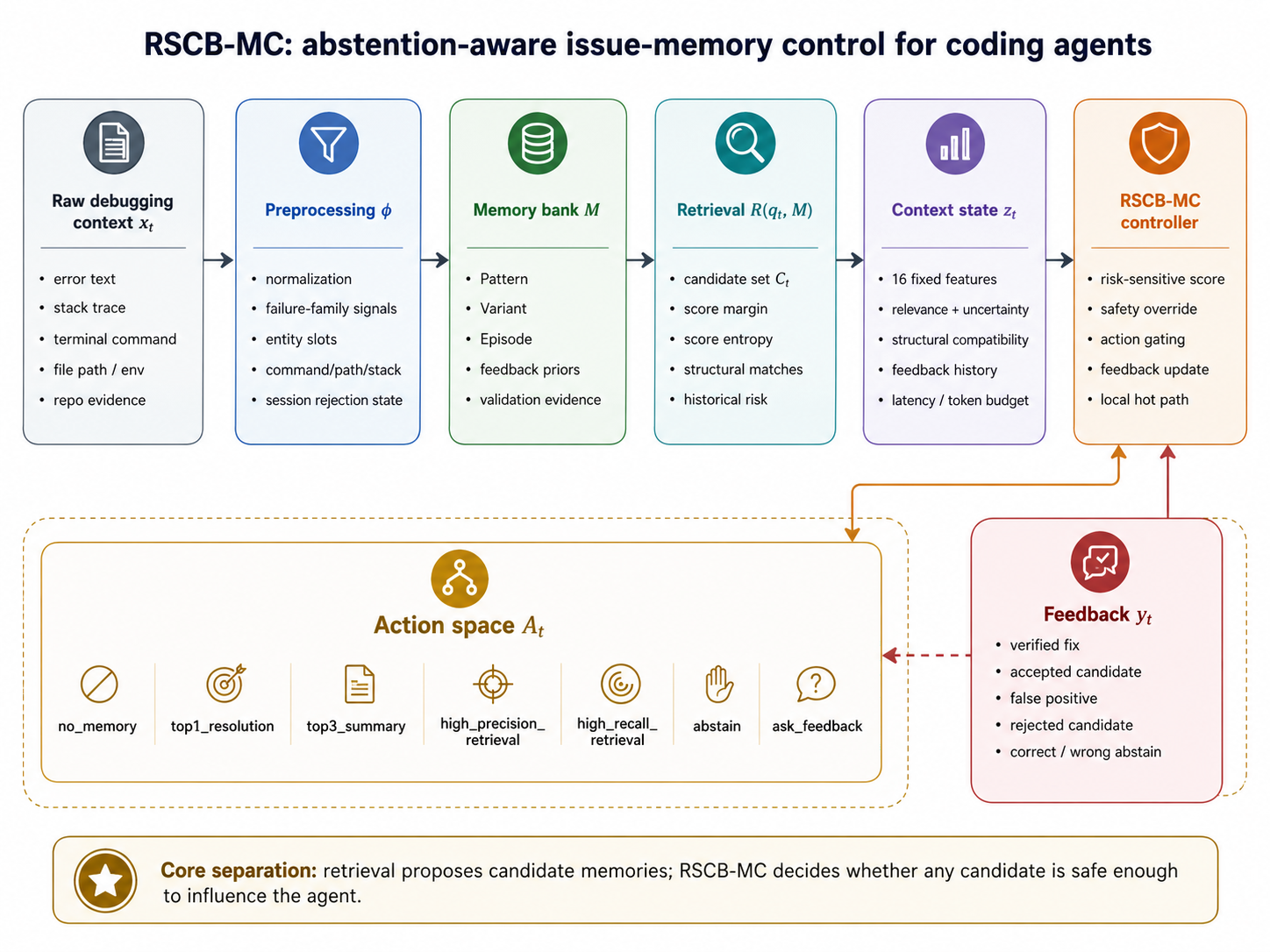}
\caption{Detailed RSCB-MC pipeline. The retrieval stage proposes candidate memories, but the controller separately decides whether any candidate is safe enough to influence the coding agent. Feedback updates the policy after each local decision.}
\label{fig:framework}
\end{figure}

Figure~\ref{fig:framework} summarizes the end-to-end RSCB-MC pipeline and highlights the architectural separation that motivates this paper. On the left, a raw debugging context $x_t$ produced by the coding agent is normalized into a query profile $q_t=\phi(x_t)$, and a local memory bank $M$ is queried by the retrieval adapter $R$ to produce a candidate set $C_t$. The pipeline does not stop at this point. Instead, the candidate set and the query profile are jointly mapped into the fixed 16-feature state $z_t=\psi(q_t,C_t)$, which exposes relevance, ambiguity, structural compatibility, feedback history, false-positive risk, and operational cost. The controller then chooses an action $a_t\in A_t$ over the full action set, including non-injection actions such as \texttt{no\_memory} and \texttt{abstain}. Only after this decision is memory allowed to alter the agent's repair trajectory. The diagram emphasizes a key invariant of the system: retrieval evidence flows into the policy as input, not as a permission signal, and feedback $y_t$ closes the loop by updating $\Theta$ so that future decisions become more risk-aware.

\begin{figure}[t]
\centering
\includegraphics[width=0.95\textwidth]{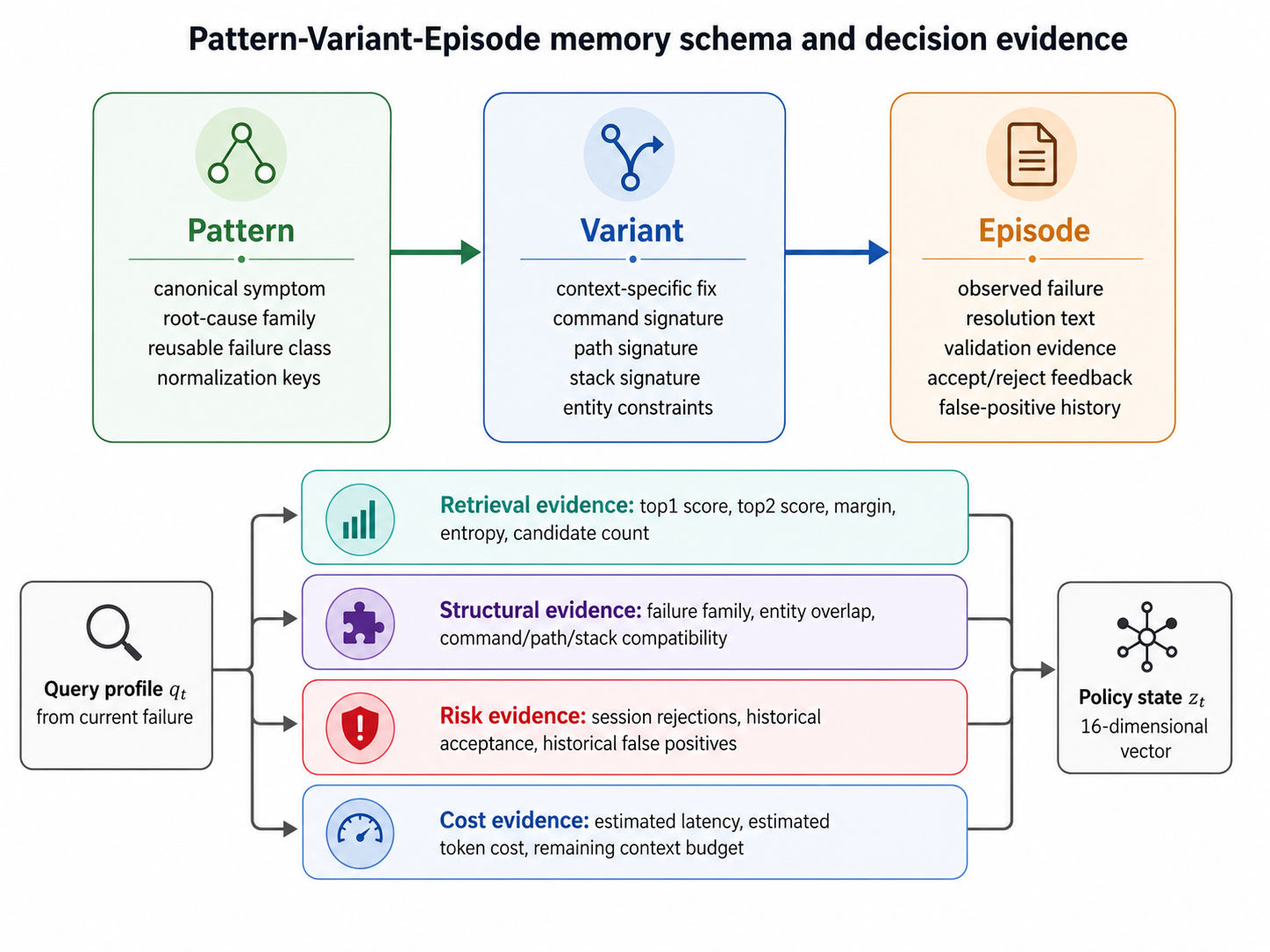}
\caption{Pattern-variant-episode memory schema. The schema separates canonical root-cause patterns, context-specific variants, and validated episodes, then exposes retrieval, structural, risk, and cost evidence to the policy state.}
\label{fig:schema}
\end{figure}

Figure~\ref{fig:schema} unfolds the three-level memory record $m_i=(p_i,V_i,E_i)$ used by RSCB-MC and shows why this layering is essential for safe reuse in coding agents. The pattern layer $p_i$ captures a canonical symptom and root-cause class that is intentionally repository-agnostic, so that the same lesson can be reused across debugging sessions. The variant layer $V_i$ refines that pattern with context-specific fix strategies and operational signatures over commands, paths, and stack traces, which lets the controller distinguish failures that share surface symptoms but differ in underlying cause. The episode layer $E_i$ stores concrete observed evidence, validation outcomes, and feedback for each historical case, which provides the auditable history needed for risk estimation. The right side of the figure makes the controller-facing view explicit: each layer projects different evidence into the 16-feature state---retrieval scores from candidate matching, structural compatibility from variants and episodes, risk evidence from the historical false-positive rate, and cost evidence from estimated payload size and latency. This structured projection is what enables the controller to reason about whether a retrieved memory is genuinely compatible with the current failure rather than merely lexically similar.

\subsection{Candidate evidence and uncertainty features}\label{subsec:evidence}

Each candidate $c_{t,i}$ receives a compatibility score $s_{t,i}\geq 0$ derived from available lexical, structural, failure-family, entity, command, path, stack, and historical feedback evidence. We write this abstractly as
\begin{equation}
    s_{t,i}=F(q_t,c_{t,i}),
\end{equation}
where $F$ denotes the deterministic candidate scoring and feature aggregation used by the local retrieval adapter. The controller does not assume that $s_{t,i}$ is a calibrated probability. Instead, it extracts uncertainty statistics from the score distribution. Let
\begin{equation}
    \pi_{t,i}=\frac{s_{t,i}+\epsilon}{\sum_{j=1}^{k}(s_{t,j}+\epsilon)}
\end{equation}
for a small smoothing constant $\epsilon>0$. Candidate entropy is
\begin{equation}
    H_t=-\sum_{i=1}^{k}\pi_{t,i}\log \pi_{t,i},
\end{equation}
and the score margin is
\begin{equation}
    \Delta_t=s_{t,(1)}-s_{t,(2)},
\end{equation}
where $s_{t,(1)}$ and $s_{t,(2)}$ are the top two candidate scores. A large margin with low entropy suggests a concentrated retrieval result; a small margin or high entropy suggests ambiguity. Ambiguity is not automatically failure, but it increases the burden on the controller before memory can be safely injected.

\subsection{The 16-feature contextual state}\label{subsec:state}

The policy state $z_t$ has fixed order and fixed dimension. Table~\ref{tab:features} lists the implemented features. The features intentionally mix relevance, ambiguity, structural compatibility, feedback history, and operational cost. This is the main difference between a top-$k$ retriever and a memory controller: the retriever describes what was found, while the controller reasons about whether that evidence is safe, useful, and affordable.

\begin{table}[t]
\caption{RSCB-MC state vector $z_t\in\mathbb{R}^{16}$.}
\label{tab:features}
\scriptsize
\begin{tabularx}{\textwidth}{@{}llX@{}}
\toprule
Index & Feature & Interpretation \\
\midrule
1 & \texttt{top1\_score} & Best candidate compatibility score. \\
2 & \texttt{top2\_score} & Second-best candidate compatibility score. \\
3 & \texttt{score\_margin} & Difference between the top two scores. \\
4 & \texttt{candidate\_entropy} & Ambiguity of the candidate score distribution. \\
5 & \texttt{candidate\_count} & Number of retrieved candidates. \\
6 & \texttt{family\_confidence} & Confidence that candidates match the same failure family. \\
7 & \texttt{entity\_match\_ratio} & Overlap between query and memory entities. \\
8 & \texttt{command\_signature\_match} & Compatibility of command-level evidence. \\
9 & \texttt{path\_signature\_match} & Compatibility of path-level evidence. \\
10 & \texttt{stack\_signature\_match} & Compatibility of stack-level evidence. \\
11 & \texttt{session\_rejection\_count} & Recent session-level memory rejections. \\
12 & \texttt{historical\_acceptance\_rate} & Prior acceptance rate for similar contexts. \\
13 & \texttt{historical\_false\_positive\_rate} & Prior false-positive rate for similar contexts. \\
14 & \texttt{estimated\_latency\_ms} & Estimated decision or payload latency. \\
15 & \texttt{estimated\_token\_cost} & Estimated memory payload token cost. \\
16 & \texttt{token\_budget\_remaining} & Remaining context-budget allowance. \\
\botrule
\end{tabularx}
\end{table}

\subsection{Action space}\label{subsec:actions}

The implemented action space is
\begin{equation}
\begin{aligned}
A=\{&\texttt{no\_memory},\texttt{top1\_resolution},\texttt{top3\_summary},\texttt{high\_precision\_retrieval},\\
&\texttt{high\_recall\_retrieval},\texttt{abstain},\texttt{ask\_feedback}\}.
\end{aligned}
\end{equation}
The action semantics are summarized in Table~\ref{tab:actions}. The important point is that abstention and no-memory are legitimate outcomes of a memory controller. They are not missing predictions; they are safety-preserving actions.

\begin{table}[t]
\caption{Memory-control actions.}
\label{tab:actions}
\begin{tabularx}{\textwidth}{@{}lXc@{}}
\toprule
Action & Meaning & Injects memory? \\
\midrule
\texttt{no\_memory} & Continue the agent trajectory without external issue memory. & No \\
\texttt{top1\_resolution} & Inject the highest-scoring resolution. & Yes \\
\texttt{top3\_summary} & Inject a compact summary of the top three candidates. & Yes \\
\texttt{high\_precision\_retrieval} & Use only high-specificity evidence. & Yes \\
\texttt{high\_recall\_retrieval} & Retrieve broader evidence when missing a valid memory is costly. & Yes \\
\texttt{abstain} & Explicitly decline memory reuse under unsafe evidence. & No \\
\texttt{ask\_feedback} & Ask for clarification or external feedback before reuse. & No \\
\botrule
\end{tabularx}
\end{table}

\subsection{Reward design}\label{subsec:reward}

Feedback is mapped to a decomposed reward
\begin{equation}
\begin{split}
 r_t = {} & \alpha I_{\mathrm{verified}} + \beta I_{\mathrm{accepted}} + \kappa I_{\mathrm{correct\_abstain}} \\
 & - \gamma I_{\mathrm{false\_positive}} - \delta I_{\mathrm{rejected}} - \eta L_t - \lambda T_t,
\end{split}
\label{eq:reward}
\end{equation}
where $L_t$ is latency cost and $T_t$ is token cost. The default coefficients are
\begin{equation}
(\alpha,\beta,\kappa,\gamma,\delta,\eta,\lambda)=(2.0,1.0,0.5,4.0,1.0,0.001,0.01).
\end{equation}
The ordering
\begin{equation}
    \gamma>\alpha>\beta
\end{equation}
is a design invariant. A false-positive memory injection is more harmful than a successful accepted reuse is beneficial. This mirrors the operational reality of debugging: a wrong fix can move the repository farther from repair, while a missed memory can still leave the agent free to reason from current evidence.

Let $\mathcal{I}\subset A$ denote injection actions. A false-positive memory injection event is
\begin{equation}
    \mathrm{FP}_t = \mathbb{1}\{a_t\in\mathcal{I}\}\,\mathbb{1}\{\text{memory is judged unsafe or causally wrong}\}.
\end{equation}
In the reported deterministic artifacts, this judgment is provided by replay/proxy metadata rather than by live user feedback.

\subsection{Risk-sensitive action score}\label{subsec:score}

The controller scores each available action through a risk-regularized objective
\begin{equation}
    S(a,z_t)=\widehat{R}(a,z_t)-\mu\widehat{p}_{fp}(a,z_t)-\lambda_c\widehat{c}(a,z_t),
\label{eq:basic_score}
\end{equation}
where $\widehat{R}$ is expected reward, $\widehat{p}_{fp}$ is estimated false-positive probability, and $\widehat{c}$ is a normalized cost. The implemented policy augments this score with action-level quality, adoption evidence, confidence terms, and contextual reward/risk models:
\begin{equation}
\begin{split}
 S_t(a) = {} & Q_t(a)+\rho A_t(a)-\mu FP_t(a)-\lambda_c C(a) \\
& + w_u\sqrt{\frac{\log(N_t+2)}{N_t(a)+1}}
- w_r\sqrt{\frac{\log(N_t+2)}{N^{fp}_t(a)+1}}\,\widehat{p}^{ctx}_{fp}(z_t) \\
& + g_\theta(a,z_t)-h_\theta(a,z_t).
\end{split}
\label{eq:full_score}
\end{equation}
Here $Q_t(a)$ is empirical action quality, $A_t(a)$ is adoption evidence, $FP_t(a)$ is action-level false-positive evidence, $C(a)$ is cost, $N_t(a)$ is the action feedback count, $N_t$ is the total feedback count, and $N^{fp}_t(a)$ is false-positive evidence for action $a$. The terms $g_\theta$ and $h_\theta$ are lightweight contextual reward and risk models over the 16-feature vector. The selected action is
\begin{equation}
    a_t=\arg\max_{a\in A_t} S_t(a),
\end{equation}
subject to safety override rules.

\begin{figure}[t]
\centering
\includegraphics[width=0.94\textwidth]{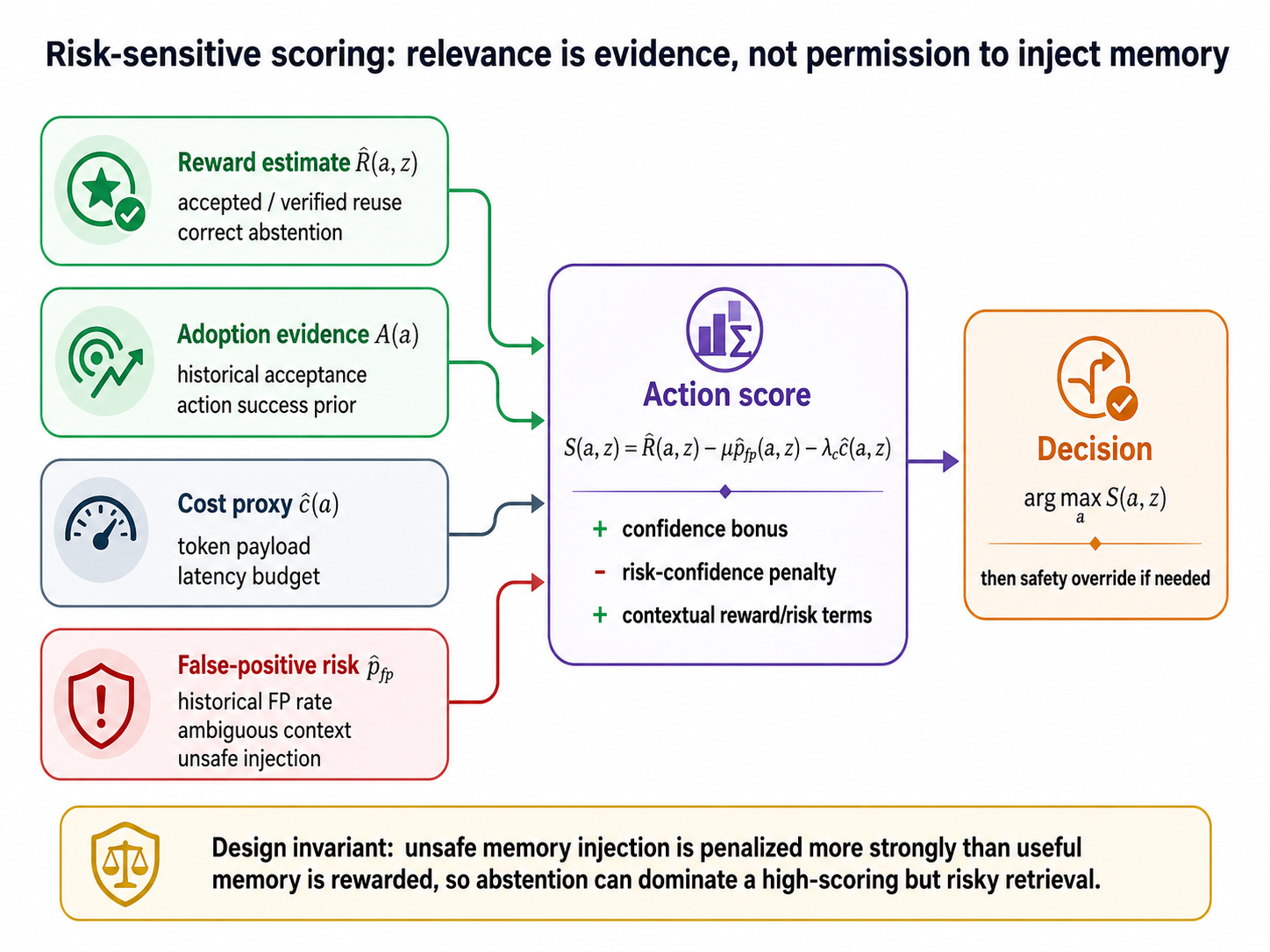}
\caption{Risk-sensitive scoring rule. RSCB-MC separates reward evidence from false-positive risk and cost; a high retrieval score is not sufficient for injection when the false-positive term dominates.}
\label{fig:score}
\end{figure}

Figure~\ref{fig:score} visualizes how the action score $S_t(a)$ in Eq.~(\ref{eq:full_score}) combines heterogeneous evidence into a single risk-aware decision. The reward branch on the left aggregates empirical action quality $Q_t(a)$, adoption evidence $A_t(a)$, an exploration bonus that scales with $\sqrt{\log(N_t+2)/(N_t(a)+1)}$, and the contextual reward model $g_\theta(a,z_t)$. The risk branch on the right aggregates the action-level false-positive term $\mu FP_t(a)$, a context-conditional penalty that scales the false-positive exploration term by $\widehat{p}^{ctx}_{fp}(z_t)$, the contextual risk model $h_\theta(a,z_t)$, and the cost term $\lambda_c C(a)$. The two branches are subtracted, which makes the scoring rule directly express the design invariant $\gamma>\alpha>\beta$: a candidate with a high retrieval score is not sufficient for injection when the historical false-positive evidence, the contextual risk model, or the cost penalty dominates the reward terms. The figure also shows the safety-override stage applied after the $\arg\max$, which can promote \texttt{abstain} or \texttt{no\_memory} when session-level rejection signals or hard-negative cues indicate that injection is unsafe regardless of the raw score. This separation between scoring and override is what allows RSCB-MC to keep the false-positive rate at zero on the hard-negative and hot-path validation sets while still selecting useful injections elsewhere.

\subsection{When refusing to remember is the right move}\label{subsec:propositions}

The statements below are not theorems in any deep sense; they are direct algebraic consequences of the reward function in Eq.~\ref{eq:reward} and the score in Eq.~\ref{eq:basic_score}. We label them \emph{decision boundaries} rather than propositions to make that scope explicit.

Let $a_I\in\mathcal{I}$ be an injection action and $a_A$ be a non-injection action such as abstain or no-memory. Let $p_v$, $p_a$, $p_f$, and $p_r$ denote the probabilities that $a_I$ produces verified reuse, accepted reuse, false-positive injection, and rejection. Let $p_c$ denote the probability that $a_A$ is a correct abstention. Ignoring small latency and token terms for readability, the expected rewards satisfy
\begin{equation}
\mathbb{E}[r\mid a_I]=\alpha p_v+\beta p_a-\gamma p_f-\delta p_r,
\end{equation}
\begin{equation}
\mathbb{E}[r\mid a_A]=\kappa p_c.
\end{equation}

\paragraph{Decision boundary 1 (when non-injection dominates).}
For a fixed context $z_t$, a non-injection action $a_A$ has greater expected reward than an injection action $a_I$ exactly when
\begin{equation}
    \gamma p_f + \delta p_r + \kappa p_c > \alpha p_v + \beta p_a,
\label{eq:dominance}
\end{equation}
because subtracting $\mathbb{E}[r\mid a_I]$ from $\mathbb{E}[r\mid a_A]$ leaves
$\kappa p_c-\alpha p_v-\beta p_a+\gamma p_f+\delta p_r$,
and the inequality is preserved when $a_A$ has no larger token or latency cost than $a_I$ under the cost terms of Eq.~\ref{eq:reward}. The point of stating Eq.~\ref{eq:dominance} is operational: under the design invariant $\gamma>\alpha>\beta$, even a moderate false-positive probability $p_f$ can flip the boundary in favour of $a_A$, which is precisely the regime we observed on hard negatives.

\paragraph{Decision boundary 2 (the score as a one-step Lagrangian).}
The action score in Eq.~\ref{eq:basic_score} is the one-step Lagrangian relaxation of
\begin{equation}
\max_{a\in A_t} \widehat{R}(a,z_t)\quad\text{s.t.}\quad \widehat{p}_{fp}(a,z_t)\leq \epsilon,\quad \widehat{c}(a,z_t)\leq b,
\end{equation}
with Lagrangian
$\mathcal{L}(a)=\widehat{R}(a,z_t)-\mu(\widehat{p}_{fp}(a,z_t)-\epsilon)-\lambda_c(\widehat{c}(a,z_t)-b)$.
For fixed $\epsilon$ and $b$, the constants $\mu\epsilon+\lambda_c b$ do not affect the maximizing action, leaving exactly Eq.~\ref{eq:basic_score}. This is a re-reading of the score, not an optimality claim: it justifies treating $\mu$ and $\lambda_c$ as risk-budget and cost-budget multipliers rather than free hyperparameters.

Neither boundary asserts global optimality of the policy over a full debugging trajectory. They only clarify the local control logic: when false-positive risk is sufficiently high, abstention or no-memory is not conservative noise; under the reward model, it is the rational action.

\subsection{Algorithm}\label{subsec:algorithm}

Algorithm~\ref{alg:rscb} summarizes RSCB-MC. The benchmark implementation is local-first and deterministic: it does not require hosted LLM calls, external embedding services, or network access.

\begin{algorithm}[t]
\caption{RSCB-MC memory control}
\label{alg:rscb}
\begin{algorithmic}[1]
\Require Raw debugging context $x_t$, memory bank $M$, policy state $\Theta$.
\State $q_t \gets \phi(x_t)$ \Comment{normalize error, command, path, stack, entity, and family evidence}
\State $C_t \gets R(q_t,M)$ \Comment{retrieve candidate memories}
\State $z_t \gets \psi(q_t,C_t)$ \Comment{build 16-feature state vector}
\State $A_t \gets \mathrm{AvailableActions}(C_t,\Theta)$
\For{each $a\in A_t$}
    \State Estimate $\widehat{R}(a,z_t)$, $\widehat{p}_{fp}(a,z_t)$, and $\widehat{c}(a,z_t)$
    \State Compute $S_t(a)$ using Eq.~\ref{eq:full_score}
\EndFor
\State $a_t \gets \arg\max_{a\in A_t} S_t(a)$
\If{$\mathrm{SafetyOverride}(z_t,C_t,\Theta)$ is true}
    \State Replace unsafe injection with \texttt{abstain} or \texttt{no\_memory}
\EndIf
\State $o_t \gets \mathrm{ApplyAction}(a_t,C_t)$
\State Observe feedback $y_t$
\State $r_t \gets \mathrm{ComputeReward}(y_t,o_t)$ using Eq.~\ref{eq:reward}
\State $\Theta \gets \mathrm{Update}(\Theta,a_t,z_t,r_t,y_t)$
\State \Return selected action $a_t$, memory output $o_t$, updated policy state $\Theta$
\end{algorithmic}
\end{algorithm}

\section{Experiments and Results}\label{sec:experiments}

\subsection{Evaluation boundary}\label{subsec:boundary}

The evaluation uses the supplied \texttt{pythalab-codex-issue-memory} artifact, package root \texttt{codex\_issue\_memory}, branch \texttt{main}, commit \texttt{227bbe3}, and Python 3.12.3. All reported results in this manuscript are generated from \texttt{data/paper\_seed}. The scripts run locally and deterministically. They do not call a hosted LLM, external embedding service, or network API. The larger \texttt{data/paper\_benchmark} dataset is present as a full-scale target, but the results reported here are smoke-scale/proxy evidence unless explicitly stated otherwise.

This boundary is deliberate. The paper evaluates the memory-control mechanism, not the full repair quality of an arbitrary production LLM agent. Offline replay and bounded hot-path validation are useful for testing safety, latency, and policy behavior, but they are not causal online counterfactual experiments.

\subsection{Benchmark artifacts}\label{subsec:artifacts}

Table~\ref{tab:data} summarizes the current smoke-scale artifacts and the full target scale. The failure families include duplicate server instances, SQLite initialization and locking, wrong virtual environment paths, wrong \texttt{PYTHONPATH}, lock-file conflict, stale migration, retrieval false positive, session-level rejected memory reuse, wrong environment variable, outdated memory variant, migration-order mismatch, corrupted local state, missing backup directory, invalid configuration key, and runtime diagnostic evidence gaps.

\begin{table}[t]
\caption{Benchmark scale. Reported results use the smoke-scale artifacts.}
\label{tab:data}
\begin{tabular*}{\textwidth}{@{\extracolsep\fill}lrr}
\toprule
Artifact & Smoke scale & Full target \\
\midrule
Canonical retrieval queries & 24 & 400 \\
Paraphrase variants & 96 & 1600 \\
Hard-negative cases & 32 & 1000 \\
Feedback replay events & 40 per seed & 3000 \\
Memory records & 16 & 80 \\
Memory variants & 32 & 160 \\
Context-budget proxy cases & 24 & 400 \\
\botrule
\end{tabular*}
\end{table}

\subsection{Baselines and metrics}\label{subsec:baselines}

The retrieval baselines are \texttt{lexical\_only}, \texttt{static\_hybrid}, \texttt{static\_hybrid\_with\_abstention}, \texttt{full\_system}, and \texttt{oracle\_upper\_bound}. The RL/bandit baselines are \texttt{epsilon\_greedy}, \texttt{ucb1}, \texttt{thompson}, \texttt{linucb}, \texttt{risk\_sensitive\_thompson}, \texttt{full\_rscb\_mc}, and \texttt{oracle\_upper\_bound}. The oracle is a non-deployable diagnostic upper bound.

The evaluation reports retrieval metrics (Recall@1, Recall@3, MRR, nDCG@3, top-1 accuracy), safety metrics (false-positive rate, unsafe injection rate, correct abstention rate, wrong abstention rate), replay metrics (average reward, cumulative reward, success rate, regret proxy), context metrics (token cost, latency proxy, risk-adjusted utility), and hot-path metrics (mean and p95 decision latency). All rates in the tables below are reported as percentages.

\subsection{Retrieval and paraphrase behavior}\label{subsec:retrieval_results}

On the canonical smoke-scale retrieval set, all retrieval methods obtain 100.0\% Recall@1, 100.0\% Recall@3, 100.0\% MRR, 100.0\% nDCG@3, and 100.0\% top-1 accuracy over 24 queries. This result should not be overinterpreted: it shows that the canonical seed set is easy for deterministic matching. The paraphrase set is more informative. Table~\ref{tab:paraphrase} shows that lexical-only and static-hybrid retrieval degrade under paraphrasing, while the oracle remains at 100.0\%. RSCB-MC is not a raw retrieval baseline in this experiment; as a controller over full-system candidates, it is evaluated through downstream memory-use behavior rather than direct Recall@1.

\begin{table}[t]
\caption{Paraphrase robustness. Percentages are computed over 24 original and 96 paraphrase queries.}
\label{tab:paraphrase}
\footnotesize
\begin{tabular*}{\textwidth}{@{\extracolsep\fill}lrrrr}
\toprule
Method & Original R@1 & Paraphrase R@1 & Original MRR & Paraphrase MRR \\
\midrule
Lexical & 100.0\% & 78.1\% & 100.0\% & 84.0\% \\
Static hybrid & 100.0\% & 80.2\% & 100.0\% & 85.1\% \\
Static+abstention & 100.0\% & 51.0\% & 100.0\% & 52.1\% \\
Full retriever & 100.0\% & 51.0\% & 100.0\% & 52.1\% \\
Oracle upper bound & 100.0\% & 100.0\% & 100.0\% & 100.0\% \\
\botrule
\end{tabular*}
\end{table}

The paraphrase result matters because it separates two problems. Canonical matching can be saturated, while robust memory-use control remains difficult. A safe memory controller should not assume that high canonical retrieval accuracy eliminates the need for abstention, structural matching, or false-positive risk estimation.

\subsection{Hard-negative safety and abstention}\label{subsec:hard_negative}

Hard negatives are cases that look similar to stored memories but should not be injected. Table~\ref{tab:hardnegative} shows the main safety behavior. Static hybrid retrieval injects memory on all hard negatives and reaches a 75.0\% false-positive rate. Risk-sensitive Thompson and full RSCB-MC produce 0.0\% false-positive rate and 100.0\% correct safe non-injection behavior on these cases.

\begin{table}[t]
\caption{Hard-negative safety over 32 cases.}
\label{tab:hardnegative}
\footnotesize
\begin{tabular*}{\textwidth}{@{\extracolsep\fill}lrrrr}
\toprule
Method & False-positive & Unsafe injection & Correct safe non-injection & Cases \\
\midrule
Static hybrid & 75.0\% & 100.0\% & 0.0\% & 32 \\
Static+abstention & 0.0\% & 0.0\% & 100.0\% & 32 \\
Thompson & 50.0\% & 50.0\% & 31.2\% & 32 \\
LinUCB & 75.0\% & 100.0\% & 0.0\% & 32 \\
Risk-sensitive TS & 0.0\% & 0.0\% & 100.0\% & 32 \\
RSCB-MC & 0.0\% & 0.0\% & 100.0\% & 32 \\
Oracle upper bound & 0.0\% & 0.0\% & 100.0\% & 32 \\
\botrule
\end{tabular*}
\end{table}

The abstention benchmark contains 56 cases. Under the conservative threshold used in the smoke-scale artifacts, full RSCB-MC and risk-sensitive Thompson answer 0.0\% of cases, incur 0.0\% risk, correctly abstain on 57.1\%, and wrongly abstain on 42.9\%. This reveals both the benefit and the cost of the current safety setting. The controller prevents unsafe injection but is over-conservative on some answerable cases. The correct future direction is calibration improvement, not removal of abstention.

\subsection{Offline replay algorithm comparison}\label{subsec:algorithm_comparison}

Table~\ref{tab:algorithms} summarizes the main offline replay comparison across two deterministic seeds and 40 replay events per seed. The success metric counts accepted/verified reuse or safe abstention according to replay/proxy labels. Full RSCB-MC has the best non-oracle success rate, 62.5\%, and a 0.0\% false-positive rate. The oracle upper bound reaches 67.5\% success, showing a remaining gap of 5.0 percentage points.

\begin{figure}[t]
\centering
\includegraphics[width=0.98\textwidth]{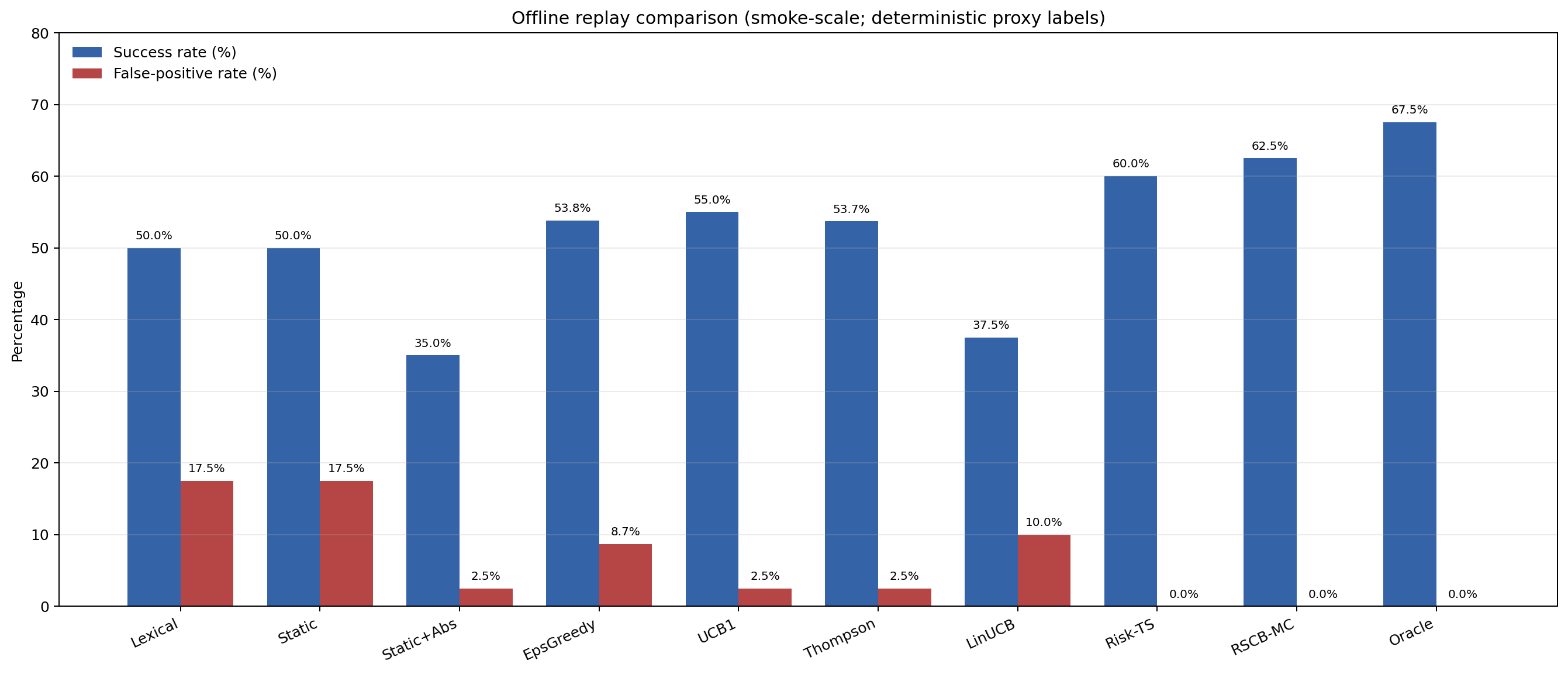}
\caption{Offline replay comparison. RSCB-MC provides the strongest non-oracle success rate while keeping the false-positive rate at 0.0\%. The oracle is non-deployable and included only as a diagnostic upper bound.}
\label{fig:alg_compare}
\end{figure}

\begin{table}[t]
\caption{Offline replay comparison. Rates are percentages; rewards are deterministic proxy rewards.}
\label{tab:algorithms}
\small
\begin{tabular*}{\textwidth}{@{\extracolsep\fill}lrrrrr}
\toprule
Method & Success & False-positive & Abstention & Verified reuse & Cum. reward \\
\midrule
Lexical & 50.0\% & 17.5\% & 0.0\% & 25.0\% & -70.800 \\
Static & 50.0\% & 17.5\% & 0.0\% & 25.0\% & -70.800 \\
Static+Abs & 35.0\% & 2.5\% & 75.0\% & 10.0\% & -33.350 \\
Epsilon-greedy & 53.8\% & 8.7\% & 30.0\% & 21.3\% & -49.818 \\
UCB1 & 55.0\% & 2.5\% & 50.0\% & 17.5\% & -35.865 \\
Thompson & 53.7\% & 2.5\% & 38.8\% & 20.0\% & -52.315 \\
LinUCB & 37.5\% & 10.0\% & 50.0\% & 15.0\% & -76.775 \\
Risk-TS & 60.0\% & 0.0\% & 37.5\% & 22.5\% & -33.700 \\
RSCB-MC & \textbf{62.5\%} & \textbf{0.0\%} & 35.0\% & \textbf{25.0\%} & \textbf{-31.740} \\
Oracle & 67.5\% & 0.0\% & 50.0\% & 25.0\% & -5.900 \\
\botrule
\end{tabular*}
\end{table}

The important comparison is not only RSCB-MC versus static retrieval. Static retrieval has 50.0\% success but a 17.5\% false-positive rate. RSCB-MC improves success to 62.5\% while reducing false positives to 0.0\%. This is the intended behavior of risk-sensitive memory control: it should not merely retrieve more memories; it should prevent dangerous memories from becoming prompt-level influence.

\subsection{Ablations}\label{subsec:ablations}

\begin{figure}[t]
\centering
\includegraphics[width=0.90\textwidth]{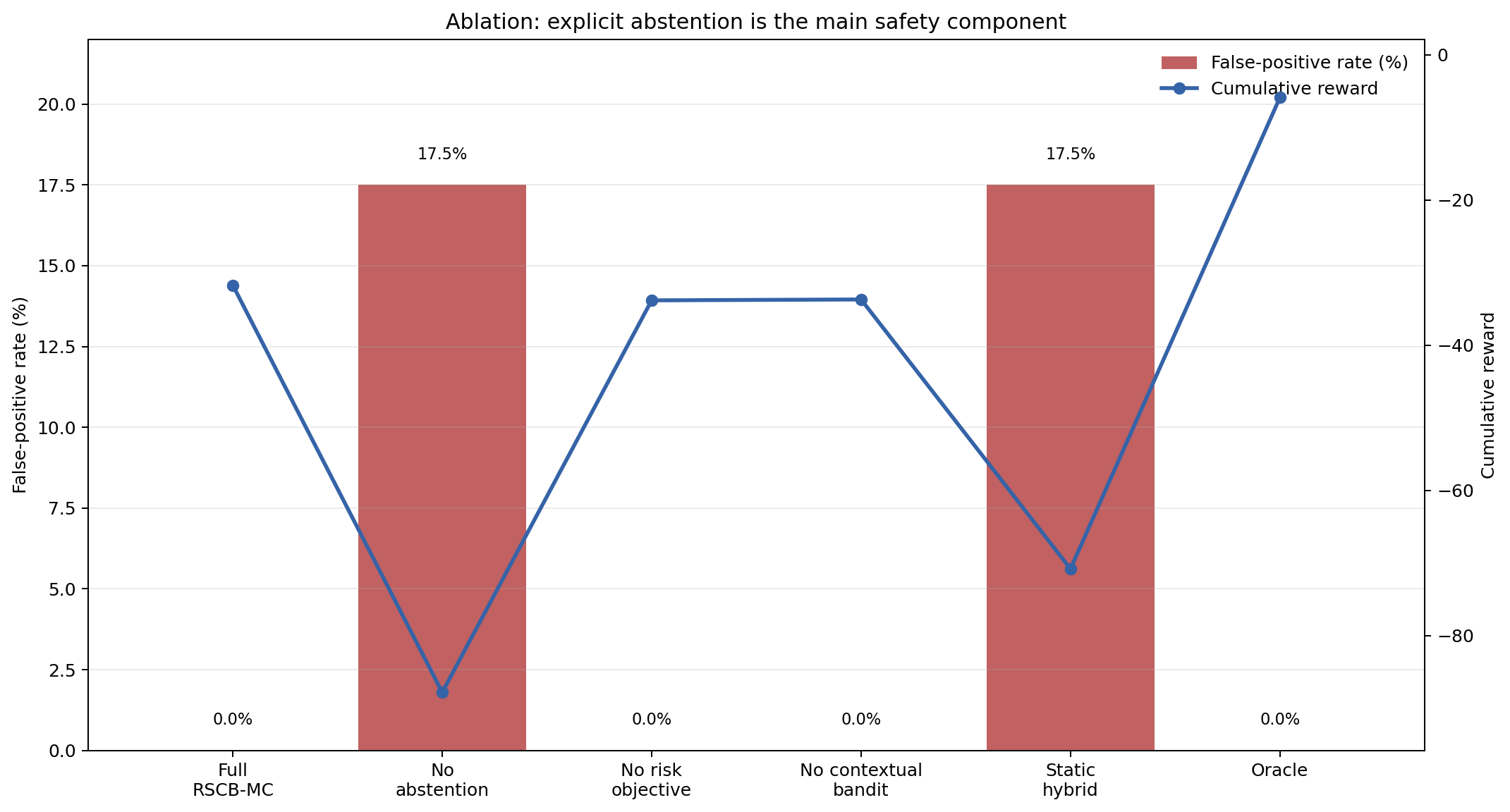}
\caption{Ablation result. Removing abstention produces the largest safety degradation, increasing false-positive rate to 17.5\% and sharply reducing cumulative reward.}
\label{fig:ablation}
\end{figure}

The strongest ablation is removal of abstention and other non-injection actions. As shown in Table~\ref{tab:ablation} and Fig.~\ref{fig:ablation}, \texttt{minus\_abstention} raises the false-positive rate from 0.0\% to 17.5\% and decreases cumulative reward from -31.740 to -87.800. This confirms that abstention is not a cosmetic fallback; it is a functional safety component.

\begin{table}[t]
\caption{Selected ablations. Percentages are rates; reward is cumulative proxy reward over replay.}
\label{tab:ablation}
\small
\begin{tabular*}{\textwidth}{@{\extracolsep\fill}lrrr}
\toprule
Variant & False-positive & Correct abstention & Cum. reward \\
\midrule
Full RSCB-MC & 0.0\% & 5.0\% & -31.740 \\
No failure-family features & 0.0\% & 5.0\% & -41.740 \\
No entity features & 0.0\% & 5.0\% & -33.495 \\
No abstention & 17.5\% & 0.0\% & -87.800 \\
No contextual bandit & 0.0\% & 5.0\% & -33.700 \\
No risk-sensitive objective & 0.0\% & 5.0\% & -33.820 \\
Static hybrid only & 17.5\% & 0.0\% & -70.800 \\
Oracle upper bound & 0.0\% & 5.0\% & -5.900 \\
\botrule
\end{tabular*}
\end{table}

Some ablations are explicitly not applicable in the current artifact because the corresponding components are not exposed by the offline replay path. We keep those cases out of the table rather than inventing unsupported measurements.

\subsection{Reward sensitivity and context budget}\label{subsec:budget}

The reward-sensitivity sweep covers false-positive penalties of -1.0, -2.5, -4.0, and -8.0, accepted-reuse rewards, correct-abstention rewards, and token penalties. The best risk-adjusted configuration in the smoke-scale sweep is \texttt{fp1\_acc0p5\_cab0p5\_tok0}, with 0.0\% false-positive rate and risk-adjusted utility -14.982. The experiment also reveals an evaluation limitation: reward magnitude changes alter reported utility directly, but in the current implementation they may not always alter action selection unless they change the signs or categories consumed by the policy update.

\begin{figure}[t]
\centering
\includegraphics[width=0.92\textwidth]{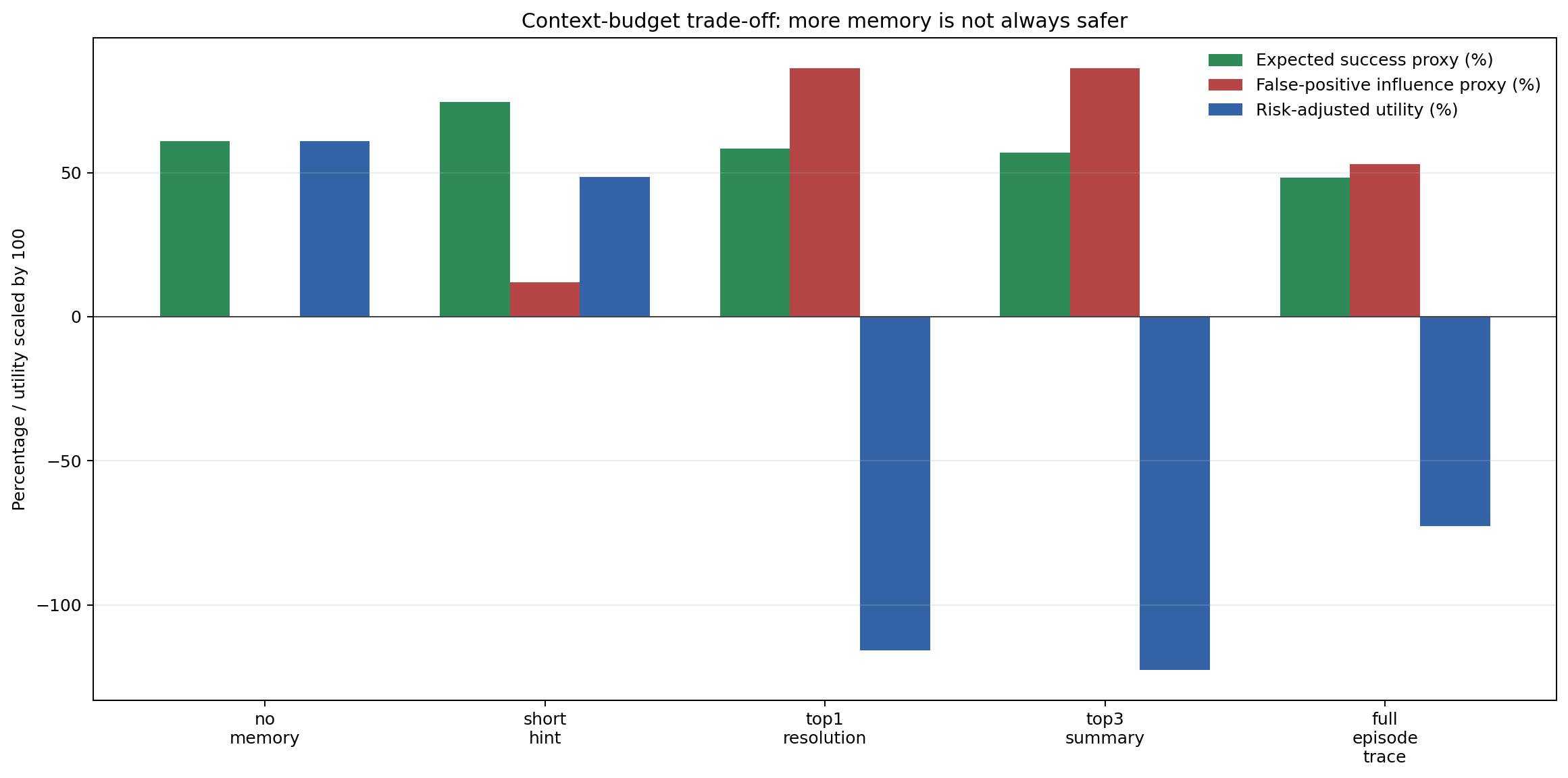}
\caption{Context-budget proxy. The short hint improves expected success, but risk-adjusted utility penalizes false-positive influence and payload cost.}
\label{fig:context}
\end{figure}

The context-budget proxy in Table~\ref{tab:context} and Fig.~\ref{fig:context} supports a simple conclusion: more memory is not always better. \texttt{short\_hint} has the highest expected success proxy, 74.5\%, but it also has a 12.0\% false-positive influence proxy. \texttt{no\_memory} has lower expected success, 60.8\%, but the best risk-adjusted utility because it has no token cost, no latency cost, and no false-positive influence. Full traces are expensive and not dominant under the current proxy.

\begin{table}[t]
\caption{Context-budget proxy over 24 cases. Percentages are proxy rates; token and latency are estimates.}
\label{tab:context}
\small
\begin{tabular*}{\textwidth}{@{\extracolsep\fill}lrrrrr}
\toprule
Mode & Tokens & Latency & Exp. success & FP influence & Utility \\
\midrule
No memory & 0.0 & 0.0 ms & 60.8\% & 0.0\% & 60.8\% \\
Short hint & 41.1 & 17.8 ms & 74.5\% & 12.0\% & 48.6\% \\
Top-1 resolution & 29.2 & 15.9 ms & 58.3\% & 86.3\% & -115.9\% \\
Top-3 summary & 157.4 & 54.3 ms & 56.8\% & 86.3\% & -122.7\% \\
Full trace & 351.1 & 114.6 ms & 48.3\% & 53.0\% & -72.7\% \\
\botrule
\end{tabular*}
\end{table}

\subsection{Bounded live hot-path validation}\label{subsec:live}

The live hot-path validation executes 200 in-process decisions per method, for 2000 total decisions. It does not call a live LLM; reward labels remain deterministic replay proxies. Table~\ref{tab:live} reports success, false-positive rate, and decision latency. Full RSCB-MC reaches 60.5\% success with 0.0\% false positives. Its p95 decision latency is 331.466 $\mu$s, which is sub-millisecond and lower than the risk-sensitive Thompson p95 latency of 411.821 $\mu$s. LinUCB is substantially slower in this hot-path setting, with p95 latency of 3567.384 $\mu$s.

\begin{figure}[t]
\centering
\includegraphics[width=0.88\textwidth]{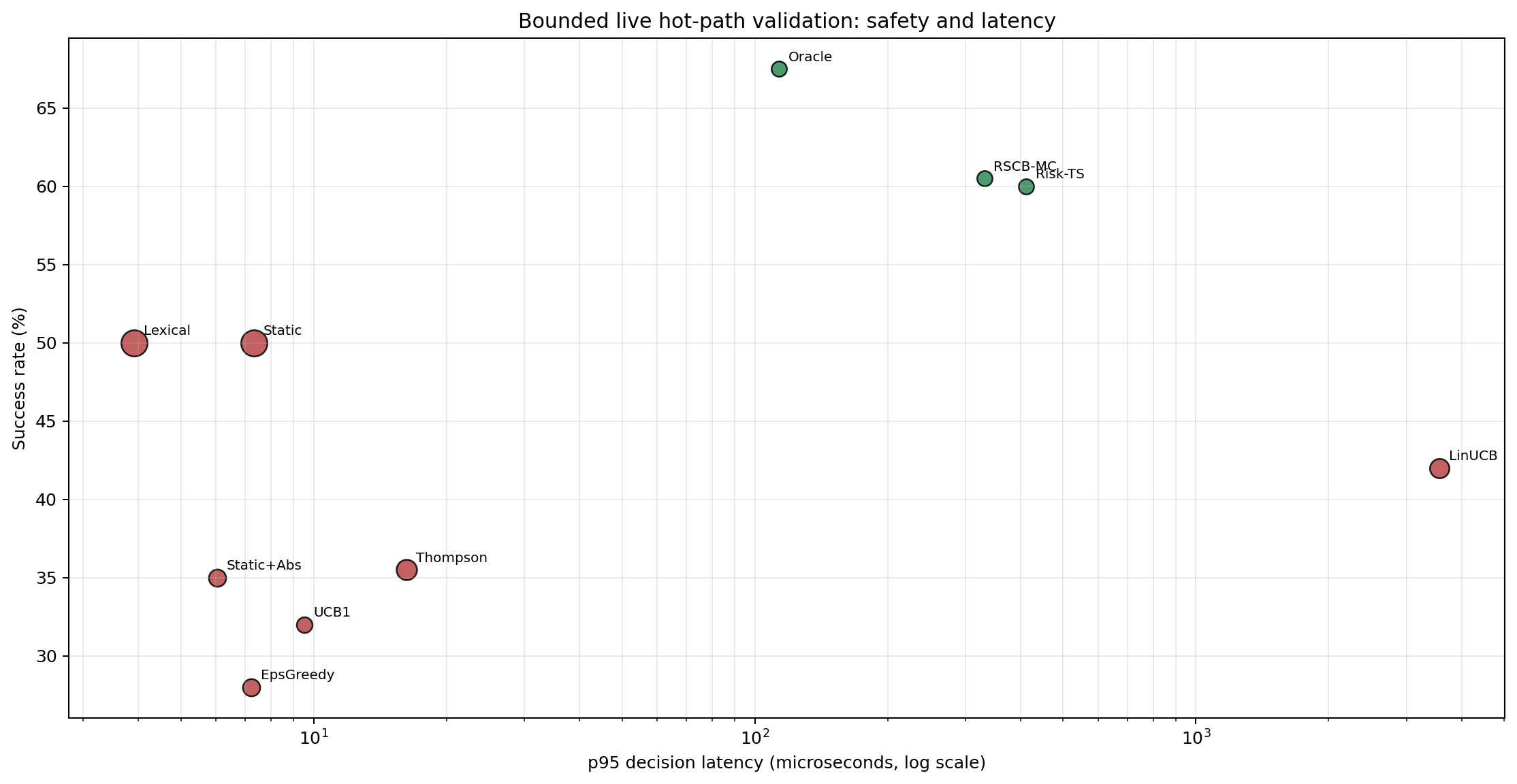}
\caption{Bounded hot-path validation. RSCB-MC maintains 0.0\% false positives with sub-millisecond p95 decision latency. Points with nonzero false-positive rates are shown in red.}
\label{fig:hotpath}
\end{figure}

\begin{table}[t]
\caption{Bounded 200-case hot-path validation. Rates are percentages.}
\label{tab:live}
\small
\begin{tabular*}{\textwidth}{@{\extracolsep\fill}lrrrr}
\toprule
Method & Success & False-positive & Mean lat. ($\mu$s) & p95 lat. ($\mu$s) \\
\midrule
Lexical & 50.0\% & 17.5\% & 3.273 & 3.920 \\
Static & 50.0\% & 17.5\% & 6.137 & 7.327 \\
Static+Abs & 35.0\% & 2.5\% & 5.282 & 6.041 \\
Epsilon-greedy & 28.0\% & 2.5\% & 5.396 & 7.213 \\
UCB1 & 32.0\% & 0.5\% & 7.687 & 9.510 \\
Thompson & 35.5\% & 7.0\% & 14.666 & 16.230 \\
LinUCB & 42.0\% & 5.5\% & 2865.263 & 3567.384 \\
Risk-TS & 60.0\% & 0.0\% & 285.689 & 411.821 \\
RSCB-MC & \textbf{60.5\%} & \textbf{0.0\%} & 267.792 & \textbf{331.466} \\
Oracle & 67.5\% & 0.0\% & 78.106 & 113.588 \\
\botrule
\end{tabular*}
\end{table}

\section{Discussion}\label{sec:discussion}

\subsection{Reading the numbers}\label{subsec:discussion_main}

The results support a specific and bounded claim. In the supplied deterministic benchmark, canonical retrieval is not the bottleneck: it reaches 100.0\% on the easy canonical set. The bottleneck is safe use of memory under ambiguity, paraphrase shift, and hard negatives. RSCB-MC improves the non-oracle offline replay success rate to 62.5\% while keeping false-positive memory injection at 0.0\%. The hard-negative result is even more direct: static hybrid retrieval injects unsafe memory in all hard-negative cases, whereas RSCB-MC avoids false-positive injection entirely. The strongest ablation confirms the mechanism. Removing abstention raises the false-positive rate to 17.5\% and reduces cumulative reward to -87.800.

This does not mean the controller is fully calibrated. In the abstention benchmark, the conservative setting refuses all cases, causing wrong abstentions on answerable examples. The safety profile is therefore asymmetric: the current controller is better at avoiding unsafe memory than at maximizing coverage. This is acceptable for an early safety-oriented memory controller, but it is not the final target. The next technical step is not to remove abstention; it is to calibrate the boundary between safe reuse and unnecessary refusal.

\subsection{How RSCB-MC differs from RAG and knowledge-boundary methods}\label{subsec:discussion_rag}

Rather than rephrasing the related work, it is more useful to lay out the differences side by side. Table~\ref{tab:rag_compare} contrasts the closest RAG-side methods with RSCB-MC along the dimensions that matter for coding agents: what is being gated, what counts as a failure, and what the cost of a wrong decision actually is.

\begin{table}[t]
\caption{Where RSCB-MC sits relative to the closest RAG and knowledge-boundary methods.}
\label{tab:rag_compare}
\footnotesize
\begin{tabularx}{\textwidth}{@{}lXXX@{}}
\toprule
Method & Gated object & First-class failure & Cost of a wrong decision \\
\midrule
Knowledge-boundary RAG \cite{m2-9} & Whether to retrieve at all & Unhelpful retrieval & Wrong text answer \\
Parametric RAG \cite{m2-6} & How to integrate retrieved context & Reasoning degradation & Wrong text answer + token cost \\
KnowledGPT / DRO \cite{m2-7,m2-4} & Knowledge access and selection & Poor knowledge selection & Wrong text answer \\
Calibration-oriented RAG \cite{m7-15} & Confidence in retrieved evidence & Mis-calibrated decision & Wrong downstream choice \\
RSCB-MC (this work) & Whether retrieved memory may enter the agent's context & False-positive memory injection in a debugging trajectory & Wrong file edits, repeated failed commands, wasted context budget \\
\botrule
\end{tabularx}
\end{table}

The philosophical stance is the same---retrieval score alone is not a permission---but the failure surface differs. In general RAG, a bad passage produces a bad answer; in coding-agent memory, a bad memory produces operational damage that can persist across turns. That is why RSCB-MC exposes command, path, stack, and session-rejection signals to the controller: in the failure modes that motivated the system, those structural signals were the only ones that distinguished, e.g., a stale-migration episode from a database-lock episode after the surface error message had already collided.

\subsection{A critical view of repair-ingredient retrieval}\label{subsec:discussion_apr}

Program-repair systems have repeatedly demonstrated that prior fixes are useful---history-driven repair learns from recurring human fix patterns \cite{m4-7}, RAP-Gen retrieves bug-fix pairs to guide patch generation \cite{m4-8}, ReAPR injects retrieved bug-fix pairs into LLM repair prompts \cite{m4-1}, and Repair-Ingredients-Search separates root-cause ingredients from solution-phase ingredients \cite{m4-2}. RSCB-MC is not a competitor to these systems; it is a safety layer that should sit \emph{in front} of them. The points below summarize where the existing repair-retrieval design and our memory-control design pull in different directions:

\begin{itemize}
  \item \textbf{Default action.} Repair-retrieval systems default to ``inject the best retrieved fix'' once a similar past case is found. RSCB-MC's default is ``do not inject unless evidence beyond similarity supports it.'' On the hard-negative set, this single difference moved the false-positive rate from 75.0\% (static hybrid) to 0.0\%.
  \item \textbf{Failure asymmetry.} Repair-retrieval rewards typically count successful patches. They do not separately measure how often a retrieved ingredient produced a wrong patch that an APR pipeline then had to revert. Our reward design makes that cost explicit through $\gamma > \alpha > \beta$.
  \item \textbf{Granularity of memory.} Repair-retrieval usually treats a retrieved bug-fix pair as a single unit. The pattern-variant-episode schema instead separates the symptom from the operational signature, which is what allowed us to distinguish, e.g., ``wrong virtual environment'' from ``wrong \texttt{PYTHONPATH}'' even when both produced the same surface trace.
  \item \textbf{Scope of evaluation.} Repair-retrieval results are usually reported on Defects4J or GitBug-Java. We make no such claim. RSCB-MC is not a patch generator and would be misjudged on a patch-success benchmark.
\end{itemize}

The net effect is complementary rather than competitive: a production repair pipeline can keep its retrieval and patch-generation components and add RSCB-MC as a gate that decides whether the retrieved ingredient is allowed to influence the next agent step at all.

\subsection{An honest trade-off against process-supervised agentic RAG}\label{subsec:discussion_agentic_rag}

It would be misleading to present RSCB-MC as strictly superior to process-supervised agentic RAG. RAG-Gym \cite{m7-9}, ReasonRAG \cite{m7-8}, TreePS-RAG \cite{m7-6}, and SIRAG \cite{m7-7} all operate at a higher level of expressivity: they optimize multi-step retrieval and reasoning trajectories, can assign credit across long chains, and can in principle learn behaviors that a single-step controller cannot represent. RSCB-MC cannot do any of that. It sees one decision at a time, scores a fixed action set, and updates with a deterministic feedback signal.

What RSCB-MC offers in exchange is a different operating point. It does not require an annotated rollout tree, a process-reward model, a critic, or a hosted LLM in the training loop. It runs in-process, makes a decision in a few hundred microseconds at p95, and produces an audit trail that a human reviewer can read action-by-action. For a coding agent that must decide \emph{before every memory injection} whether the retrieved evidence is safe, that latency-and-auditability profile is the binding constraint. Process-supervised methods are the right answer when the goal is end-to-end retrieval-and-reasoning quality; RSCB-MC is the right answer when the goal is a fast, transparent gate that prevents unsafe memory from entering the agent's context in the first place. The two are best read as complementary layers, not as competing solutions to the same problem.

\subsection{Where memory-management RL ends and RSCB-MC begins}\label{subsec:discussion_memory_rl}

Memory-R1 and Agentic Memory teach agents the full lifecycle of memory operations: add, update, delete, retrieve, summarize, discard \cite{m7-12,m7-14}. MIRIX further organizes this lifecycle across multiple specialized memory types in a multi-agent architecture \cite{m3-4}. RSCB-MC does much less than any of these. It does not write memory, does not compress memory off-line, does not resolve contradictions between competing entries, and does not decide when an old episode should be retired. It only answers one question at read time: ``is this retrieved memory safe to inject right now?''

This is a deliberate scoping choice, and it has a real downside. RSCB-MC inherits whatever memory the upstream write path produces, including stale, duplicated, or contradictory entries. If a wrong fix has been written into memory and re-validated by an unreliable feedback signal, RSCB-MC will eventually trust it. The right way to read this paper is therefore not as a replacement for memory-management RL but as the read-time safety layer that such systems are currently missing. In a complete production stack, Memory-R1-style learning would manage what enters memory, and an RSCB-MC-style controller would decide whether what comes out of memory is allowed to alter the agent's next move. Memory aging, contradiction resolution, and privacy governance, all of which appear in memory surveys as open problems \cite{m3-1}, remain firmly outside our scope.

\subsection{What changes when abstention targets memory rather than answers}\label{subsec:discussion_abstention}

Abstention has become a respectable answer-time strategy: selective prediction \cite{m5-1}, selective generation \cite{m5-3}, conformal abstention \cite{m5-2}, learning-to-defer \cite{m5-7}, and Learn-to-Refuse \cite{m5-6} all show that refusing to answer can improve reliability under uncertainty, and recent work argues that prompt-level incentives alone can already shift LLMs toward more honest abstention \cite{m5-4,m5-8}. We borrow the core idea but move the decision boundary one step earlier in the pipeline. The controller is not deciding whether the LLM should answer; it is deciding whether a retrieved memory should be inserted into the prompt that the LLM will then read.

This displacement changes what is being measured. A wrong generated answer is visible at the output layer and can be flagged, scored, or rolled back; a wrong memory injection is hidden inside the prompt and silently biases every reasoning step that follows. Treating false-positive memory injection as its own first-class metric, separate from any final-answer accuracy, is therefore not a stylistic preference. It is the only way to see the failure mode at all. The ablation study makes this concrete: when abstention is removed and the controller is forced to commit on every ambiguous case, the false-positive rate jumps from 0.0\% to 17.5\% and cumulative reward collapses from -31.740 to -87.800. The lesson is narrow but useful: abstention literature has been studying the right idea on the wrong object for coding-agent memory.

\subsection{What this means in practice}\label{subsec:practical}

Three practical implications follow. First, memory systems for coding agents should report false-positive injection rates, not only Recall@k or MRR. Second, memory payload size should be controlled jointly with risk. The context-budget proxy suggests that short hints may be useful, but large traces and summaries can be dominated when they carry high false-positive influence or high token cost. Third, memory gating should be separated from retrieval. This separation makes the system easier to audit: one can inspect why a memory was retrieved and separately why it was or was not injected.

\subsection{Where this evidence stops}\label{subsec:limitations}

The evidence has several boundaries. First, the reported benchmark is smoke-scale. The full benchmark artifacts exist, but final journal claims should be refreshed using the full-scale pipeline. Second, the offline replay evaluation is not a causal online counterfactual guarantee. When a policy selects an action, feedback is obtained from deterministic replay metadata rather than from live user outcomes. Third, the hot-path validation measures in-process decision cost and update behavior, but its reward labels are still offline proxies. Fourth, the context-budget and agent dry-run experiments are proxy tasks, not direct measurements of production LLM repair quality. Fifth, the contextual-bandit formulation optimizes immediate memory-control decisions and does not model delayed repository evolution, multi-turn repair consequences, or non-stationary memory aging.

These limitations do not invalidate the core mechanism, but they constrain the claim. The appropriate conclusion is that RSCB-MC provides a coherent, efficient, and safety-oriented memory-control mechanism in the supplied deterministic artifacts. It does not yet prove broad production superiority across arbitrary repositories, LLMs, or live coding-agent deployments.

\section{Conclusion}\label{sec:conclusion}

This paper introduced RSCB-MC, a risk-sensitive contextual bandit controller for abstention-aware memory retrieval in LLM-based coding agents. The method reframes issue-memory use as a selective control problem: the agent should not only ask which memory is most similar, but whether any retrieved memory is safe enough to influence the debugging trajectory. RSCB-MC stores issue knowledge through a pattern-variant-episode schema, constructs a 16-feature context vector from retrieval evidence, and uses a reward function that penalizes false-positive memory injection more strongly than missed reuse.

The current deterministic smoke-scale artifacts show that canonical retrieval can be saturated while safe memory use remains difficult. Full RSCB-MC achieves the best non-oracle offline replay success rate, 62.5\%, with a 0.0\% false-positive rate. It preserves a 0.0\% false-positive rate in bounded live hot-path validation and remains sub-millisecond at p95 decision latency. The ablations show that explicit abstention is central to safety: removing abstention raises false positives to 17.5\% and substantially worsens cumulative reward.

Future work should extend the current evidence in five directions: full-scale benchmark refresh, live LLM-agent evaluation, better calibration of the safe-reuse boundary, off-policy evaluation with stronger counterfactual estimators, and longer-horizon modeling of delayed repair outcomes, stale memory, and repository drift. The broader lesson is simple: for coding agents, memory is useful only when the agent also learns when not to use it.

\section*{Declarations}

\textbf{Funding} Not applicable.

\textbf{Conflict of interest} The author declares no conflict of interest.

\textbf{Ethics approval and consent to participate} Not applicable. This study did not involve human participants, human data, or animal subjects.

\textbf{Consent for publication} Not applicable.

\textbf{Materials availability} Not applicable.

\textbf{Author contributions} M.I. conceived the study, designed and implemented the RSCB-MC method, conducted all experiments, analyzed the results, and wrote the manuscript.

\textbf{Data and code availability} The source code, deterministic local artifacts, and reproduction scripts supporting the findings of this study are openly available at \url{https://github.com/PhiniteLab/codex-issue-memory}. The reported smoke-scale results are generated from \texttt{data/paper\_seed}; the full target benchmark is stored under \texttt{data/paper\_benchmark} in the same repository.

\bibliography{references}

@article{m1-1,
  author = {Bui, Nghi D. Q.},
  title = {{Building Effective AI Coding Agents for the Terminal: Scaffolding, Harness, Context Engineering, and Lessons Learned}},
  year = {2026},
  journal = {arXiv preprint arXiv:2603.05344},
  eprint = {2603.05344},
  archivePrefix = {arXiv},
}

@article{m1-2,
  author = {Wang, J. and Ni, T. and Lee, W. B. and Zhao, Q.},
  title = {{A contemporary survey of large language model assisted program analysis}},
  year = {2025},
  journal = {arXiv preprint arXiv:2502.18474},
  eprint = {2502.18474},
  archivePrefix = {arXiv},
}

@article{m1-3,
  author = {Yang, J. and Liu, X. and Lv, W. and Deng, K. and Guo, S. and Jing, L. and Zheng, B. and others},
  title = {{From code foundation models to agents and applications: A comprehensive survey and practical guide to code intelligence}},
  year = {2025},
  journal = {arXiv preprint arXiv:2511.18538},
  eprint = {2511.18538},
  archivePrefix = {arXiv},
}

@article{m2-4,
  author = {Shi, Z. and Yan, L. and Sun, W. and Feng, Y. and Ren, P. and Ma, X. and Ren, Z. and others},
  title = {{Direct retrieval-augmented optimization: Synergizing knowledge selection and language models}},
  year = {2025},
  journal = {ACM Transactions on Information Systems},
}

@inproceedings{m2-6,
  author = {Su, W. and Tang, Y. and Ai, Q. and Yan, J. and Wang, C. and Wang, H. and Liu, Y. and others},
  title = {{Parametric retrieval augmented generation}},
  year = {2025},
  booktitle = {Proceedings of the 48th International ACM SIGIR Conference on Research and Development in Information Retrieval},
  pages = {1240-1250},
}

@article{m2-7,
  author = {Wang, X. and Yang, Q. and Qiu, Y. and Liang, J. and He, Q. and Gu, Z. and Wang, W. and others},
  title = {{KnowledGPT: Enhancing large language models with retrieval and storage access on knowledge bases}},
  year = {2023},
  journal = {arXiv preprint arXiv:2308.11761},
  eprint = {2308.11761},
  archivePrefix = {arXiv},
}

@article{m2-9,
  author = {Zhang, Z. and Wang, X. and Jiang, Y. and Chen, Z. and Mu, F. and Hu, M. and Huang, F. and others},
  title = {{Exploring Knowledge Boundaries in Large Language Models for Retrieval Judgment}},
  year = {2024},
  journal = {arXiv preprint arXiv:2411.06207},
  eprint = {2411.06207},
  archivePrefix = {arXiv},
}

@article{m3-1,
  author = {Du, P.},
  title = {{Memory for autonomous LLM agents: Mechanisms, evaluation, and emerging frontiers}},
  year = {2026},
  journal = {arXiv preprint arXiv:2603.07670},
  eprint = {2603.07670},
  archivePrefix = {arXiv},
}

@article{m3-4,
  author = {Wang, Y. and Chen, X.},
  title = {{MIRIX: Multi-agent memory system for LLM-based agents}},
  year = {2025},
  journal = {arXiv preprint arXiv:2507.07957},
  eprint = {2507.07957},
  archivePrefix = {arXiv},
}

@article{m4-1,
  author = {Liu, Z. and Du, X. and Liu, H.},
  title = {{ReAPR: Automatic program repair via retrieval-augmented large language models}},
  year = {2025},
  journal = {Software Quality Journal},
  volume = {33},
  number = {3},
  pages = {30},
}

@article{m4-2,
  author = {Zhang, J. and Huang, K. and Zhang, J. and Liu, Y. and Chen, C.},
  title = {{Repair Ingredients Are All You Need: Improving Large Language Model-Based Program Repair via Repair Ingredients Search}},
  year = {2025},
  journal = {arXiv preprint arXiv:2506.23100},
  eprint = {2506.23100},
  archivePrefix = {arXiv},
}

@inproceedings{m4-7,
  author = {Le, X. B. D. and Lo, D. and Le Goues, C.},
  title = {{History driven program repair}},
  year = {2016},
  booktitle = {2016 IEEE 23rd International Conference on Software Analysis, Evolution, and Reengineering (SANER)},
  volume = {1},
  pages = {213-224},
}

@inproceedings{m4-8,
  author = {Wang, W. and Wang, Y. and Joty, S. and Hoi, S. C. H.},
  title = {{RAP-Gen: Retrieval-augmented patch generation with CodeT5 for automatic program repair}},
  year = {2023},
  booktitle = {Proceedings of the 31st ACM Joint European Software Engineering Conference and Symposium on the Foundations of Software Engineering},
  pages = {146-158},
}

@inproceedings{m5-1,
  author = {Xin, J. and Tang, R. and Yu, Y. and Lin, J.},
  title = {{The art of abstention: Selective prediction and error regularization for natural language processing}},
  year = {2021},
  booktitle = {Proceedings of the 59th Annual Meeting of the Association for Computational Linguistics and the 11th International Joint Conference on Natural Language Processing (Volume 1: Long Papers)},
  pages = {1040-1051},
}

@article{m5-2,
  author = {Tayebati, S. and Kumar, D. and Darabi, N. and Jayasuriya, D. and Krishnan, R. and Trivedi, A. R.},
  title = {{Learning conformal abstention policies for adaptive risk management in large language and vision-language models}},
  year = {2025},
  journal = {arXiv preprint arXiv:2502.06884},
  eprint = {2502.06884},
  archivePrefix = {arXiv},
}

@article{m5-3,
  author = {Lee, M. and Kim, K. and Kim, T. and Park, S.},
  title = {{Selective generation for controllable language models}},
  year = {2024},
  journal = {Advances in Neural Information Processing Systems},
  volume = {37},
  pages = {50494-50527},
}

@inproceedings{m5-4,
  author = {Sch\"onw\"alder, E. and Falkenberg, C. and Hartmann, C. and Lehner, W.},
  title = {{Abstention is all you need}},
  year = {2025},
  booktitle = {2025 IEEE 12th International Conference on Data Science and Advanced Analytics (DSAA)},
  pages = {1-10},
}

@inproceedings{m5-6,
  author = {Cao, L.},
  title = {{Learn to refuse: Making large language models more controllable and reliable through knowledge scope limitation and refusal mechanism}},
  year = {2024},
  booktitle = {Proceedings of the 2024 Conference on Empirical Methods in Natural Language Processing},
  pages = {3628-3646},
}

@article{m5-7,
  author = {Montreuil, Y. and Yeo, S. H. and Carlier, A. and Ng, L. X. and Ooi, W. T.},
  title = {{Optimal query allocation in extractive QA with LLMs: A learning-to-defer framework with theoretical guarantees}},
  year = {2024},
  journal = {arXiv preprint arXiv:2410.15761},
  eprint = {2410.15761},
  archivePrefix = {arXiv},
}

@article{m5-8,
  author = {Zong, H. and Li, B. and Long, Y. and Chang, S. and Wu, J. and Hadfield, G. K.},
  title = {{I-CALM: Incentivizing Confidence-Aware Abstention for LLM Hallucination Mitigation}},
  year = {2026},
  journal = {arXiv preprint arXiv:2604.03904},
  eprint = {2604.03904},
  archivePrefix = {arXiv},
}

@inproceedings{m7-4,
  author = {Zhang, W. and Zhu, Y. and Lu, Y. and Demarne, M. and Wang, W. and Deng, K. and Krishnan, S. and others},
  title = {{FLAIR: Feedback learning for adaptive information retrieval}},
  year = {2025},
  booktitle = {Proceedings of the 34th ACM International Conference on Information and Knowledge Management},
  pages = {6284-6292},
}

@article{m7-6,
  author = {Zhang, T. and Li, K. and Li, J. and Li, Y. and Luo, H. and Wu, X. and Meng, H. and others},
  title = {{TreePS-RAG: Tree-based Process Supervision for Reinforcement Learning in Agentic RAG}},
  year = {2026},
  journal = {arXiv preprint arXiv:2601.06922},
  eprint = {2601.06922},
  archivePrefix = {arXiv},
}

@article{m7-7,
  author = {Wang, J. and Wu, Z. and Lu, S. and Li, Y. and Huang, X.},
  title = {{SIRAG: Towards Stable and Interpretable RAG with A Process-Supervised Multi-Agent Framework}},
  year = {2025},
  journal = {arXiv preprint arXiv:2509.18167},
  eprint = {2509.18167},
  archivePrefix = {arXiv},
}

@article{m7-8,
  author = {Zhang, W. and Li, X. and Dong, K. and Wang, Y. and Jia, P. and Li, X. and Zhao, X. and others},
  title = {{Process vs. Outcome Reward: Which is Better for Agentic RAG Reinforcement Learning}},
  year = {2025},
  journal = {arXiv preprint arXiv:2505.14069},
  eprint = {2505.14069},
  archivePrefix = {arXiv},
}

@article{m7-9,
  author = {Xiong, G. and Jin, Q. and Wang, X. and Fang, Y. and Liu, H. and Yang, Y. and Zhang, A. and others},
  title = {{RAG-Gym: Systematic optimization of language agents for retrieval-augmented generation}},
  year = {2025},
  journal = {arXiv preprint arXiv:2502.13957},
  eprint = {2502.13957},
  archivePrefix = {arXiv},
}

@article{m7-11,
  author = {Wu, Y. and Zheng, Y. and Xu, T. and Zhang, Z. and Yu, Y. and Zhu, J. and Yu, G. and others},
  title = {{ContextBudget: Budget-Aware Context Management for Long-Horizon Search Agents}},
  year = {2026},
  journal = {arXiv preprint arXiv:2604.01664},
  eprint = {2604.01664},
  archivePrefix = {arXiv},
}

@article{m7-12,
  author = {Yan, S. and Yang, X. and Huang, Z. and Nie, E. and Ding, Z. and Li, Z. and Ma, Y. and others},
  title = {{Memory-R1: Enhancing large language model agents to manage and utilize memories via reinforcement learning}},
  year = {2025},
  journal = {arXiv preprint arXiv:2508.19828},
  eprint = {2508.19828},
  archivePrefix = {arXiv},
}

@article{m7-14,
  author = {Yu, Y. and Yao, L. and Xie, Y. and Tan, Q. and Feng, J. and Li, Y. and Wu, L.},
  title = {{Agentic memory: Learning unified long-term and short-term memory management for large language model agents}},
  year = {2026},
  journal = {arXiv preprint arXiv:2601.01885},
  eprint = {2601.01885},
  archivePrefix = {arXiv},
}

@article{m7-15,
  author = {Jang, C. and Cho, D. and Lee, S. and Lee, H. and Lee, J.},
  title = {{Reliable Decision Making via Calibration Oriented Retrieval Augmented Generation}},
  year = {2024},
  journal = {arXiv preprint arXiv:2411.08891},
  eprint = {2411.08891},
  archivePrefix = {arXiv},
}

@article{m8-6,
  author = {Jiang, Z. and Lo, D. and Liu, Z.},
  title = {{Agentic Software Issue Resolution with Large Language Models: A Survey}},
  year = {2025},
  journal = {arXiv preprint arXiv:2512.22256},
  eprint = {2512.22256},
  archivePrefix = {arXiv},
}

@article{m8-8,
  author = {Guo, J. and Li, N. and Qi, J. and Yang, H. and Li, R. and Feng, Y. and Xu, M. and others},
  title = {{Empowering working memory for large language model agents}},
  year = {2023},
  journal = {arXiv preprint arXiv:2312.17259},
  eprint = {2312.17259},
  archivePrefix = {arXiv},
}

\end{document}